\documentclass{article}

\usepackage{arxiv}

\usepackage[utf8]{inputenc} 
\usepackage[T1]{fontenc}    
\usepackage{hyperref}       
\usepackage{url}            
\usepackage{booktabs}       
\usepackage{amsfonts}       
\usepackage{nicefrac}       
\usepackage{microtype}      
\usepackage{lipsum}

\usepackage{subfigure}
\usepackage{tikz}           
\usepackage{pgfplots}
\usepackage{amsmath}

\RequirePackage{natbib}

\usepackage{graphics}

\title{Random Bias Initialization Improves Quantized Training}

\author{
Xinlin Li \& Vahid Partovi Nia \thanks{Noah's Ark Lab} \\
Huawei Montreal Research Center \\
Montreal, QC H3N 1X9, Canada \\
\texttt{\{xinlin.li1,vahid.partovinia\}@huawei.com} \\
}

\begin{document}
\maketitle

\begin{abstract}
Binary neural networks improve computationally efficiency of deep models with a large margin. However, there is still a performance gap between a  successful full-precision training and binary training. We  bring some insights about why this accuracy drop exists and call for  a better understanding of binary network geometry.  We start with analyzing  full-precision neural networks  with ReLU activation and compare it with its binarized version. This comparison suggests to  initialize networks with random bias, a counter-intuitive remedy. 
\end{abstract}

\section{Introduction}
It is  common to use low-bit quantized networks such as Binary Neural Networks (BNNs)  \citep{Courbariaux_2016_BinarizedNN} to implement  deep neural networks on edge devices such as cell phones, smart wearables, etc. BNNs only keep the sign of weights and compute the sign of activations  $\{-1, +1\}$ by applying the sign function in the forward pass. In backward propagation, BNN uses Straight-Through-Estimator (STE) to estimate the backward gradient through the sign function and update on full-precision weights. The forward and backward loop of a BNN, therefore, becomes similar to the full-precision neural network with hard hyperbolic tangent \emph{htanh} activation. The htanh function is a piece-wise linear version of the nonlinear hyperbolic tangent (\emph{tanh}), and is known to be inferior in terms of accuracy compared to ReLU-like activation function. Although the analysis is based on htanh function, this conclusion equally applies to BNNs that use STE, a htanh-like, back propagation scheme. Other saturating activations like tanh and \emph{sigmoid} commonly applied in recurrent neural networks and attention-based models may benefit from this resolution as well.  Among others, \cite{sari2019study} recommends an initialization scheme for binary weights but ignores the bias term. \cite{Ramachandran_2017_SWISH} utilized automatic search techniques on searching different activation functions. Most top novel activation functions found by the searches have an asymmetric saturating regime, which is similar to ReLU.

\section{Related Works}
Initialization strategy is critical for training deep neural network. A good initialization helps to alleviate the exploding/vanishing gradient problem, as well as reduce training time and achieve lower generalization error. In terms of weight initialization, \cite{glorot2010understanding} proposed a new initialization scheme designed for sigmoid and tanh activations. \cite{he2015delving} extended this work and adapt to ReLU activation, which make it the most widely-used initialization scheme for training deep neural network with ReLU activation. However, all these initialization scheme set the bias values to zero. 

In terms of bias initialization for ReLU activated neural network, \cite{li2015cs231n} suggested a trick that initializing all bias value with a small constant value, like 0.01, to ensure that all ReLU units fire in the beginning to alleviate "dying ReLU" issue. Besides ReLU-activated model, bias initialization trick also helps recurrent neural network (RNN) training. For Long short-term memory (LSTM), \cite{gers1999learning} proposed initializing the forget gate biases to positive values and  \cite{jozefowicz2015empirical} suggest to initialize the forget gate with a unit constant value. The latter empirically showed that this bias initialization trick significantly improve the performance of the LSTM model and beat all LSTM alternatives, such as Gated Recurrent Units (GRU). \cite{billa2017improving} showed that initializing forget gate bias with small random values outperforms  the constant initialization. Similarly, the reset gate biases of GRU cell can be initialized with $-1$ for better performance. The intuition behind these RNN tricks is to ensure the BackProp gradient pass through all time-steps at initialization to better capture the long-term dependency of the data sequence. It's worth to mentioned that the forget gate of LSTM cell and the reset gate of GRU cell are similar to a linear layer followed by a sigmoid activation function. However, the understanding of relationship between bias initialization and activation function remains unexplored, which is the main contribution of this paper. 

There are different ways to look at deep neural network. One is to treat a neural network as an universal function approximator. \cite{cybenko1989approximation} proved the universal approximation theorem for neural network with sigmoid activation. Later, \cite{hornik1989multilayer} extended it for arbitrary activation functions. 

Recently, several works \citep{long2017pde} \citep{chang2018reversible} \citep{chen2018neural} proposed a new view that considering the forward propagation of Residual Network (ResNet) architecture as solving the initial value problem of ordinary differential equation with Euler's method \citep{Griffiths2010}. These works linked the neural network understanding to the numerical solution of differential equation.

Another way is looking at neural network from the hyper-plane point of view. This view generalizes the hyper-plane concept from support vector machine to neural network \citep{gulcehre2016noisy} \citep{van2016some}. \citep{montufar2014number}. We take this approach.

Activation function brings non-linearity into neural network and sigmoid function and tanh function were the popular choices before ReLU.  \cite{glorot2011deep} proposed training deep neurals network with ReLU activation, and argued that ReLU activation alleviates the vanishing gradient problem and encourages sparsity in the model. 


Since AlexNet \citep{krizhevsky2012imagenet}, almost every successful neural network architectures use ReLU activation or its variants, such as leaky ReLU \citep{maas2013rectifier}, parametric ReLU \citep{he2015delving}, ELU \citep{clevert2015fast}, etc. Although many works reported that ReLU activation outperforms the traditional saturating activation functions, the reason for its superior performance remains an open question. In this paper, we argue why ReLU performs better than other activations.

\cite{ramachandran2017searching} utilized automatic search on  different activation functions. Most top novel activation functions found by the searches have an asymmetric saturating regime, which is similar to ReLU. \cite{farhadi2019activation} adapts ReLU and sigmoid while training. To improve the performance of saturating activations, \cite{xu2016revise} proposed \emph{penalized tanh} activation, which introduces asymmetric saturating regime to tanh by inserting leaky ReLU before tanh. The penalized tanh achieves the same level of performance as ReLU-activated CNN.


\section{Full-Precision Networks}
 A typical full-precision neural network block can be described by 
\begin{equation} \label{eq:wx_bn_act}
	\begin{gathered}
		x^{i+1} = \text{ReLU}(W^{i}x^{i}+b^{i}) \\
		W^{i} \in \mathbb{R}^{m \times n}, b^{i} \in \mathbb{R}^{m}, x^{i} \in \mathbb{R}^n, x^{i+1} \in \mathbb{R}^m.
	\end{gathered}
\end{equation}
Neural networks are trained using the back-propagation algorithm. Back propagation is composed of two components i) forward pass and ii) backward propagation. In the forward pass, the loss function $\L(.)$ is evaluated on the current weights, and in backward propagation,  gradients and then  weights are updated sequentially.

\begin{figure}[t]
\centering
\fbox{\includegraphics[width=0.4\textwidth]{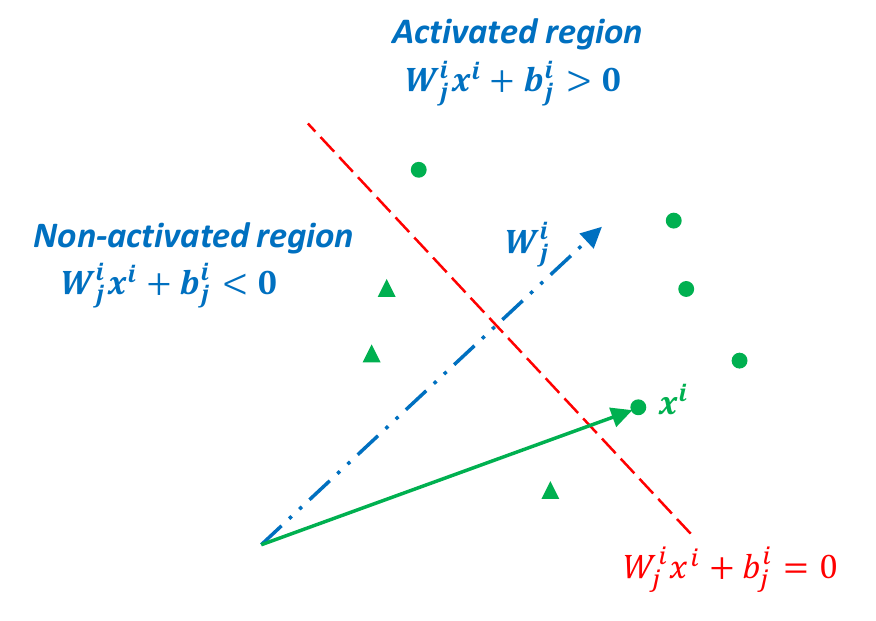}}
\fbox{\includegraphics[width=0.27 \textwidth]{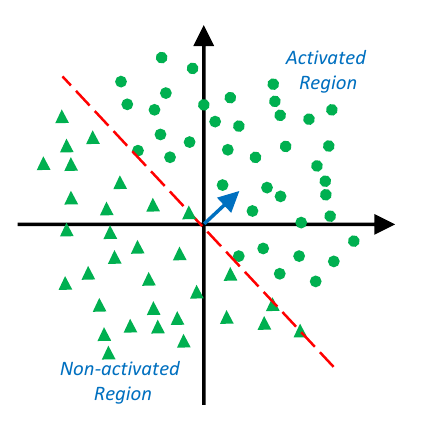}}
\caption{Activated and non-activated regions of ReLU (left panel). Activated region of ReLU at initialization (right panel).} 
\label{fig:activate_relu}
\end{figure}

\begin{figure}[t]
\centering
\fbox{\includegraphics[width=0.4\textwidth]{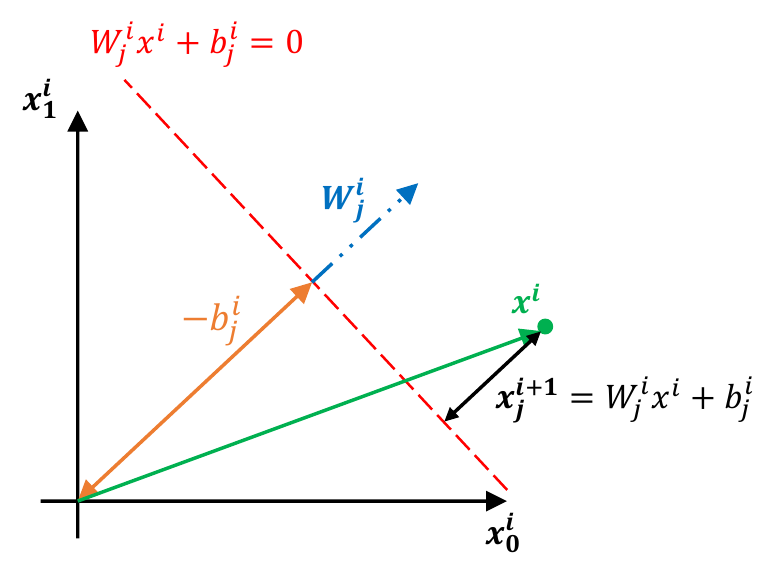}}~~\fbox{\includegraphics[width=0.33\textwidth]{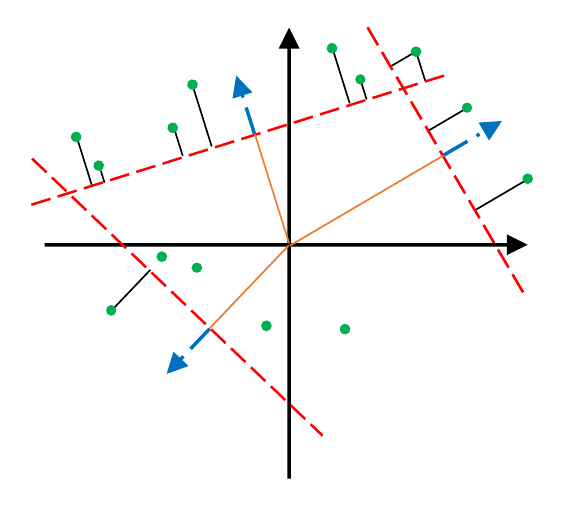}}

\caption{Geometric behavior of ReLu during forward pass, trained hyperplanes  (left panel) and their geometry (right panel).} \label{fig:hyper-planes}
\end{figure}

Assume weight vectors $W^{i}_{j}$ have unit norm. It is a reasonable assumption when the network has batch normalization layers in which all neuron responses are normalized, as the magnitude of the weight vectors does not affect the layer output.  The $j^{th}$ neuron response in the $(i+1)^{th}$ layer are computed as 
\begin{equation}
	\begin{aligned}
        x^{i+1}_{j} &= \begin{cases}
		     W^{i}_{j}x^{i}+b^{i}_{j} & W^{i}_{j}x^{i} + b^{i}_{j} > 0 \\
		     0 & W^{i}_{j}x^{i}+b^{i}_{j} \leq 0 \\
		     \end{cases}
	\end{aligned}
\end{equation}
First, the input data points $x^{i}$ are projected to the $j^{th}$ row vector of the weight matrix. The dot product of $W^{i}_{j}$ and $x^{i}$ are cut by the corresponding bias term $b^{i}_{j}$, i.e. the output $x^{i+1}_{j}$ is set to zero if the dot product is smaller than the threshold, see Figure~\ref{fig:hyper-planes} (left panel).
A hyper-plane whose normal direction defined by $W^{i}_{j}$ divides the input space into two parts: i) activated region (non-saturated regime) and ii) non-activated region (saturated regime), see Figure~\ref{fig:activate_relu}. If the data point $x^{i}$ falls on the positive side of a hyper-plane (activated region), the hyper-plane is activated by $x^{i}$. Consequently, $x^{i+1}_{j}$ is positive. Otherwise, $x^{i+1}_{j}$ equals zero and remains inactive. 

The  weight matrix $W^{i}$ of size $m\times n$
 and the bias vector $b^{i}$ of size $m \times 1 $ define $m$ hyper-planes in the $n$-dimensional input space,  see Figure~\ref{fig:hyper-planes} (right panel) .

During backward propagation, the backward gradient update on $W^{i}_{j}$ and $x^{i}$ are computed using 
\begin{equation}
\label{eq:pLoss_pXip1j}
	\begin{aligned}
	\begin{cases}
        \frac{d \L}{d W^{i}_{j}} &= 
		     \frac{d \L}{d x^{i+1}_{j}}
		     *\frac{d x^{i+1}_{j}}{d W^{i}_{j}} \\
        \frac{d \L}{d b^{i}_{j}} &= 
		     \frac{d \L}{d x^{i+1}_{j}}
		     *\frac{d x^{i+1}_{j}}{d b^{i}_{j}} \\
        \frac{d \L}{d x^{i}} &= 
		     \frac{d \L}{d x^{i+1}_{j}}
		     *\frac{d x^{i+1}_{j}}{d x^{i}}
    \end{cases}
	\end{aligned}
\end{equation}

For the case of ReLU activation
\begin{equation}
\label{eq:pXip1j_pWij}
	\begin{aligned}
        \frac{d x^{i+1}_{j}}{d W^{i}_{j}} &= \begin{cases}
		     x^{i} & W^{i}_{j}x^{i} + b^{i}_{j} > 0 \\
		     0 & W^{i}_{j}x^{i} + b^{i}_{j} \leq 0 \\
		     \end{cases}
	\end{aligned}
\end{equation}

\begin{equation}
\label{eq:pXip1j_pbij}
	\begin{aligned}
        \frac{d x^{i+1}_{j}}{d b^{i}_{j}} &= \begin{cases}
		     1 & W^{i}_{j}x^{i} + b^{i}_{j} > 0 \\
		     0 & W^{i}_{j}x^{i} + b^{i}_{j} \leq 0 \\
		     \end{cases}
	\end{aligned}
\end{equation}

\begin{equation}
\label{eq:pXip1j_pXi}
	\begin{aligned}
        \frac{d x^{i+1}_{j}}{d x^{i}} &= \begin{cases}
		     W^{i}_{j} & W^{i}_{j}x^{i} + b^{i}_{j} > 0 \\
		     0 & W^{i}_{j}x^{i} + b^{i}_{j} \leq 0 \\
		     \end{cases}
	\end{aligned}
\end{equation}
The activation function only allows the gradients from data point on the activated region to backward propagate and update the hyper-plane \eqref{eq:pXip1j_pWij}.

From the hyper-plane analysis, we realize that ReLU activation has three ideal properties i) The diversity of activated regions at initialization, ii) The equality of data points at initialization, iii) The equality of hyper-planes at initialization which we discuss each property in more details later. These may explain why ReLU activation outperforms the traditional tanh or sigmoid activations.
To argue each property, let us suppose that the distribution of the dot products is zero-centered. This assumption is automatically preserved in neural networks with batch normalization layer. 

i) Region diversity: the activated regions of hyper-planes solely depend on the direction of the weight vector, which is randomly initialized. This allows different hyper-planes to learn from a different subset of data points, and ultimately diversifies the backward gradient signal. ii) Data equality: an arbitrary data point $x^{i}$, is located on  activated regions of approximately half of the total hyper-planes in a layer. In other words, the backward gradients from all data points can pass through the approximately same amount of activation function, update hyper-planes, and propagate the gradient.  iii) Hyperplane equality:
an arbitrary hyper-plane $W^{i}_{j}$, is affected by the backward gradients from approximately $50\%$ of the total data points. All hyper-planes on average receive the same amount of backward gradients. Hyper-plane equality speeds up the convergence and facilitates model optimization, see Figure~\ref{fig:activate_relu} (right panel).

\begin{figure}
\centering
\fbox{\includegraphics[height=0.23\textheight]{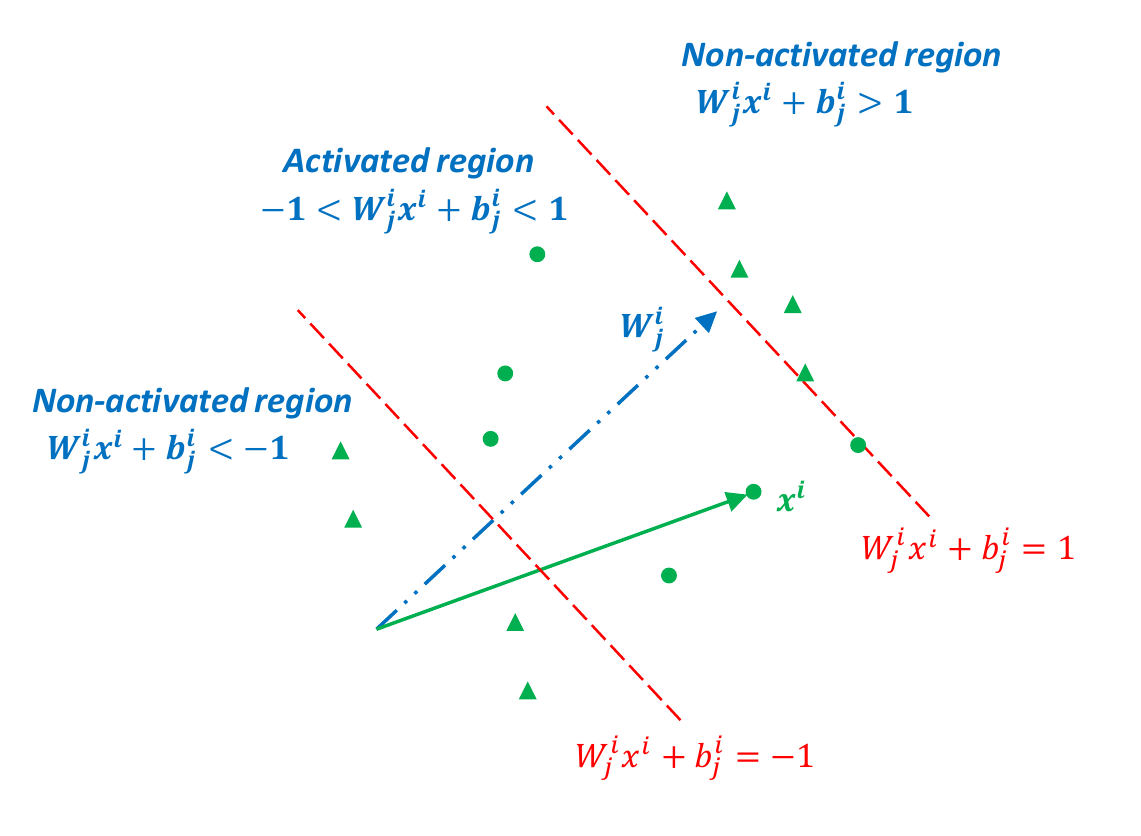}}
\fbox{\includegraphics[height=0.23\textheight]{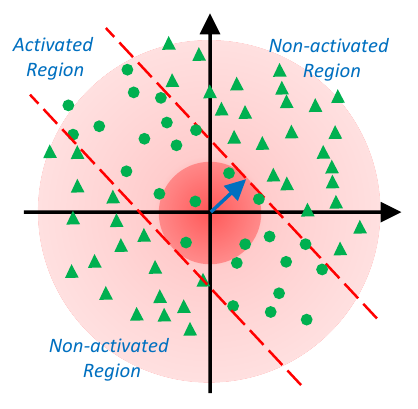}}

\caption{Activated region and non-activated region of htanh activation function(left panel). Activated region of Hard Tanh at initialization (right panel)} \label{fig:activate_htanh}
\end{figure}

The performance gap between ReLU activation and htanh activation is caused by their different activated region distribution, see Figure~\ref{fig:activate_htanh}. Clearly, htanh activation is not as good as ReLU in defining balanced and fair activated regions. However, we analyze each property for htanh as well.

i) Region diversity: activated regions of htanh are not as diverse as ReLU. Activated regions of htanh cover only the area close to the origin. Assuming Gaussian data, this is a dense area that the majority of data points are located in.

ii) Data equality: data points are not treated fairly htanh activation function. Data points that closer to the origin can activate more hyper-planes than the data points far from the origin. If the magnitude of a data point $x^{i}$ is small enough, it can activate all hyper-planes in the same layer, see the deep-red region of Figure~\ref{fig:activate_htanh} (right panel). As a consequence, in backward gradients, few data instances affect all hyper-planes. In other words, the backward gradients from a part of the training data have a larger impact on model than others. This imbalance ultimately affects model generalization problem since the model training focuses only on a subset of the training data points close to the origin.

iii) Hyperplane equality: The initial activated regions should cover a similar-sized subset of the training data points overall, and this property is shared in both ReLU and htanh activations. Similar analysis also applies to other activation functions with the zero-centered activated region, like sigmoid or tanh.

\section{Training Acceleration}
Here we proposed a simple initialization strategy to alleviate the data inequality issue and improve activated region diversity for the htanh activation relying on our geometric insight described earlier. We argue bias initialization with a uniform distribution between $[-\lambda, \lambda]$, where $\lambda$ is a hyper-parameter. With random bias initialization, the data points that far from the origin can activate more hyper-planes. If $\lambda > \text{max}(\left\lVert x\right\rVert)+1$, all data points activate approximately the same number of hyper-planes during backward propagation, so data equality can be achieved. Also, with the diverse initial activated region, different hyper-planes learn from different subset of training data points.

However, this initialization strategy comes with a drawback. Hyper-plane equality no longer holds when the biases are not set to zero. Hyper-planes with larger initial bias have less activated data. Therefore, choosing the optimal value of $\lambda$ is a trade-off between the hyper-plane equality and the data equality. Experiments below  shows that the validation curve becomes unsteady if $\lambda$ value set to too high. Empirically, with a batch normalization layer, $\lambda \approx 2$ provide a good initial estimate. In this case, the activated regions covering from $-3$ to $+3$, so it allows  the gradients from almost all data points to propagate. Our experiments shows  small $\lambda$ also helps to improve the performance of ResNet architecture.

\section{Numerical Results}
The proposed bias initialization method is evaluated on the CIFAR-10. We use the same random seed for all experiments, so they have identical weight initialization. The network architectures are based on the original implementation of the BNN  \cite{Courbariaux_2016_BinarizedNN}. We choose the VGG-7 architecture and the ResNet architecture.

\begin{figure}
\centering
\begin{subfigure}{}
\centering
\includegraphics[width=0.4\textwidth]{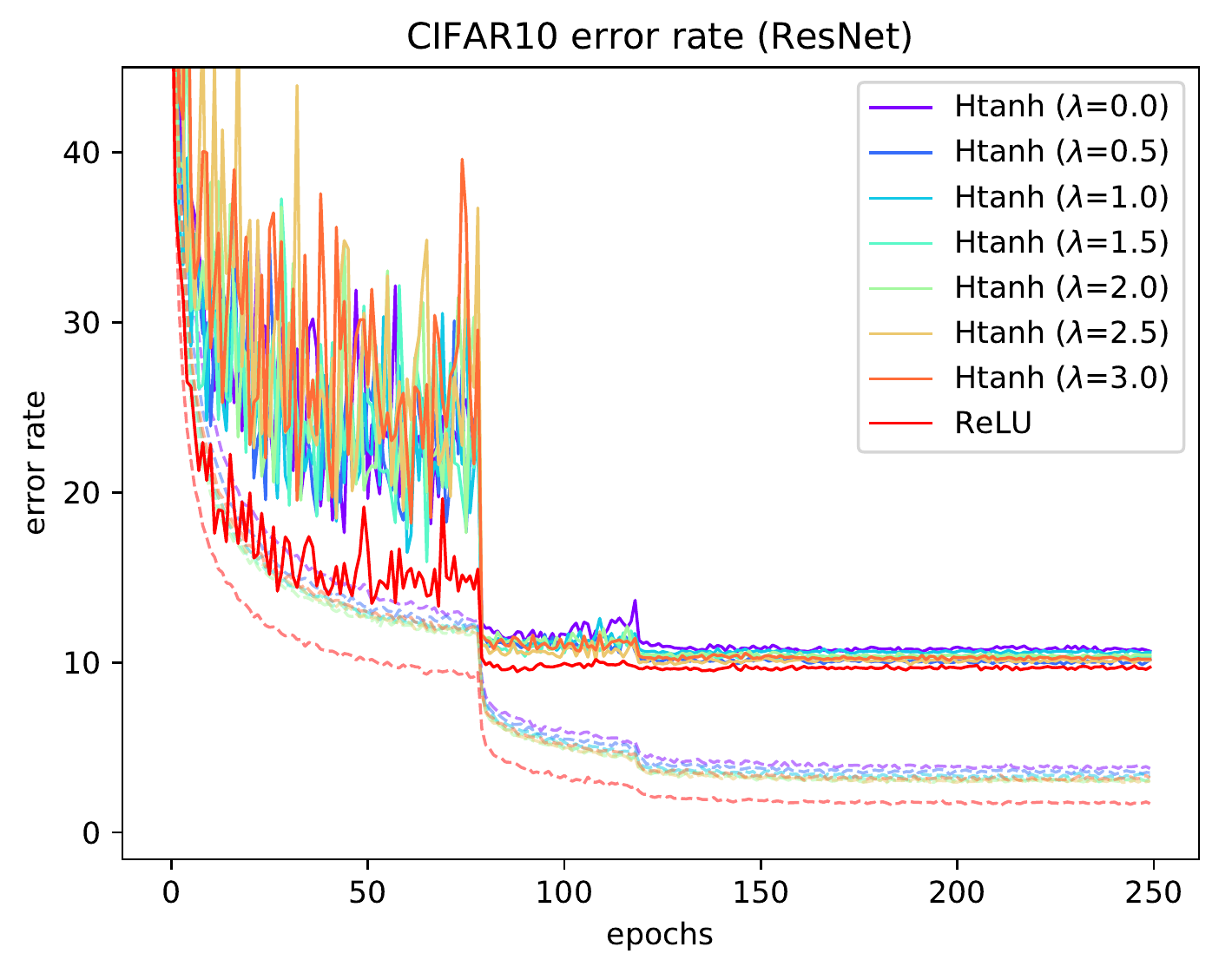}
\includegraphics[width=0.4\textwidth]{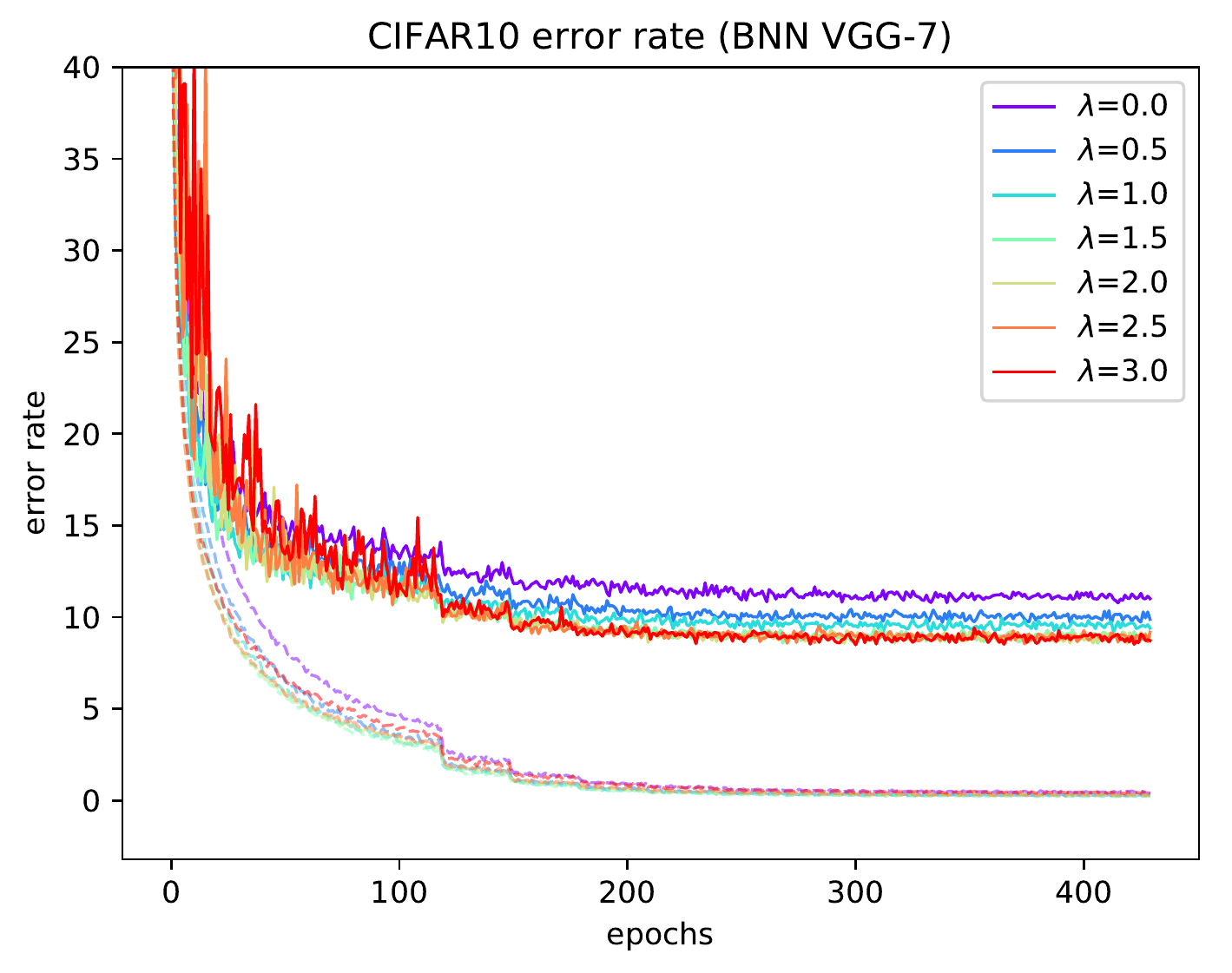}
\end{subfigure}
\begin{subfigure}{}
\includegraphics[width=0.45\textwidth]{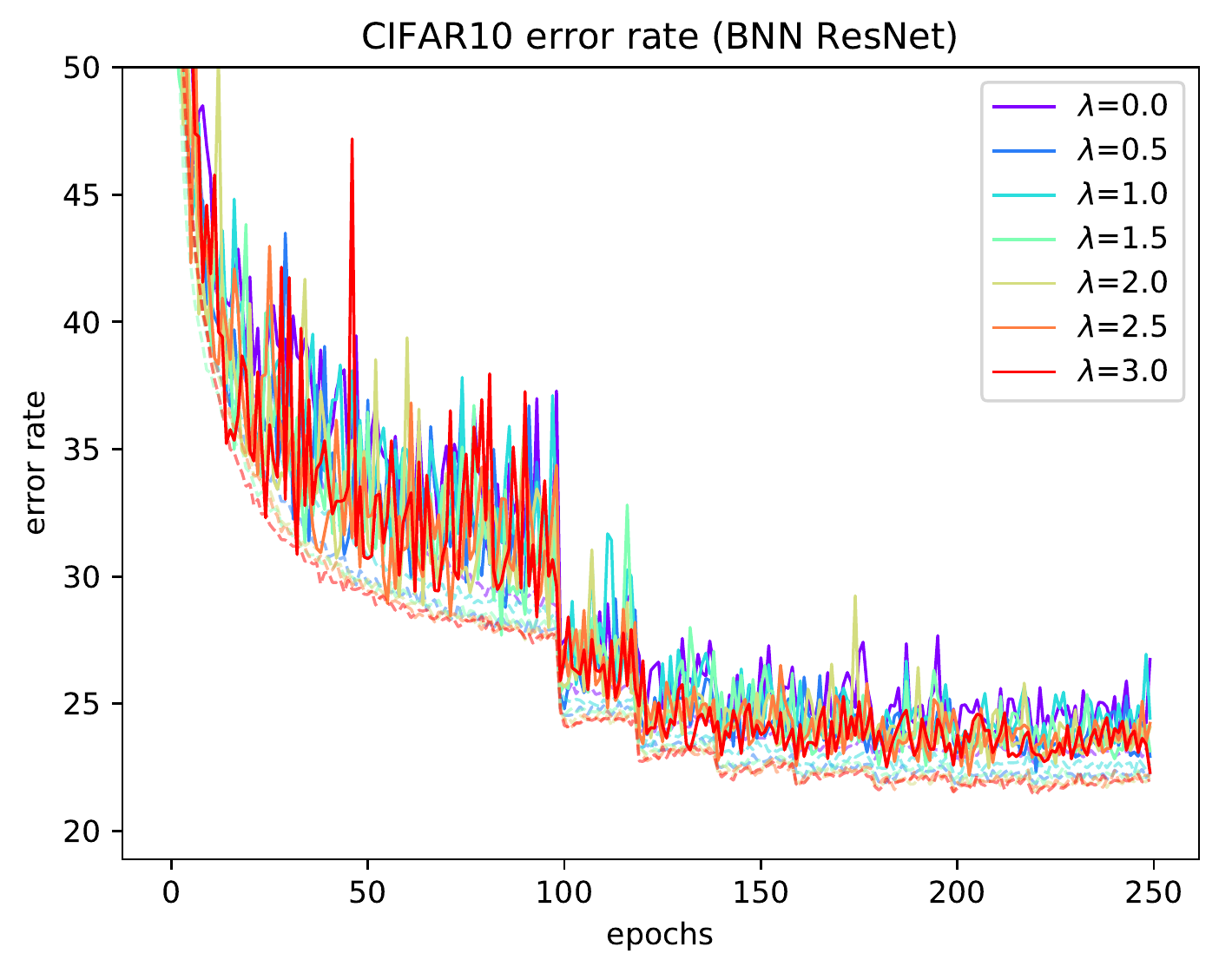}
\end{subfigure}
\caption{Training of full-precision ResNet architecture (top left panel) Binary VGG-7 architecture (top right), and Binary ResNet (bottom panel)} \label{fig:bias_init}
\end{figure}

The VGG-7 architecture, is a simple and over-parameterized model for CIFAR 10. This is an ideal architecture to compare the performance between different activations. 

Figure \ref{fig:bias_init} confirms that the random bias initialization strategy helps to reduce the performance gap between htanh and ReLU activation. A similar effect is observed for ResNet type architectures. 

\begin{table}
\begin{center}
\begin{tabular}{ c c c } 
\hline
Activations & VGG-7 & ResNet \\
\hline
ReLU (Baseline) & 6.98 & 9.45 \\ 
htanh & 10.91 & 10.63 \\ 
htanh ($\lambda$=0.5) & 9.99 & 9.87 \\ 
htanh ($\lambda$=1.0) & 9.15 & 10.47 \\ 
htanh ($\lambda$=1.5) & 8.36 & 10.13 \\ 
htanh ($\lambda$=2.0) & 7.98 & 10.23 \\ 
htanh ($\lambda$=2.5) & \textbf{7.83} & \textbf{9.84} \\ 
\hline
\end{tabular}
\end{center}
\caption{Validation error rate \% for full-precision training,  $\lambda =0 $ coincides with common deterministic initialization.}
\label{table:fp_cifar10}
\end{table}

\begin{table}
\begin{center}
\begin{tabular}{ c c c } 
\hline
$\lambda$ & Binary VGG-7 & Binary ResNet \\
\hline
0.0 & 10.77 & 23.11 \\ 
0.5 & 9.57 & 22.31 \\ 
1.0 & 9.17 & 22.83 \\ 
1.5 & 8.57 & 22.56 \\ 
2.0 & 8.56 & 22.47 \\ 
2.5 & 8.53 & \textbf{22.19} \\ 
3.0 & \textbf{8.48} & 22.30 \\ 
\hline
\end{tabular}
\end{center}
\caption{Validation error rate \% for Binary training, $\lambda =0 $ coincides with common deterministic initialization. }
\label{table:bnn_cifar10}
\end{table}



We also tested the proposed bias initialization on the ResNet-like architecture. The results are depicted in Figure~\ref{fig:bias_init} re-assures  that bias initialization improves htanh and pushes it toward ReLU accuracy, see Table \ref{table:fp_cifar10}.

Binary training that use STE is similar to htanh activation. We expect to observe a similar effect in BNN training with STE gradient approximator.  The validation error rate is summarized in Table~\ref{table:bnn_cifar10}. In the Binary VGG-7 experiments, we reduced the accuracy gap between full-precision network with ReLU activation and BNN from 4\% to 1.5\%. The bias initialization strategy is effective to close the gap on binary ResNet architecture by almost 1\%, even while the full-precision model even under-fits on CIFAR10 data.

\bibliographystyle{iclr2020}  
\bibliography{references}  

\begin{thebibliography}{25}
\providecommand{\natexlab}[1]{#1}
\providecommand{\url}[1]{\texttt{#1}}
\expandafter\ifx\csname urlstyle\endcsname\relax
  \providecommand{\doi}[1]{doi: #1}\else
  \providecommand{\doi}{doi: \begingroup \urlstyle{rm}\Url}\fi

\bibitem[Billa(2017)]{billa2017improving}
Jayadev Billa.
\newblock Improving lstm-ctc based asr performance in domains with limited
  training data.
\newblock \emph{arXiv preprint arXiv:1707.00722}, 2017.

\bibitem[Chang et~al.(2018)Chang, Meng, Haber, Ruthotto, Begert, and
  Holtham]{chang2018reversible}
Bo~Chang, Lili Meng, Eldad Haber, Lars Ruthotto, David Begert, and Elliot
  Holtham.
\newblock Reversible architectures for arbitrarily deep residual neural
  networks.
\newblock In \emph{Thirty-Second AAAI Conference on Artificial Intelligence},
  2018.

\bibitem[Chen et~al.(2018)Chen, Rubanova, Bettencourt, and
  Duvenaud]{chen2018neural}
Tian~Qi Chen, Yulia Rubanova, Jesse Bettencourt, and David~K Duvenaud.
\newblock Neural ordinary differential equations.
\newblock In \emph{Advances in neural information processing systems}, pp.\
  6571--6583, 2018.

\bibitem[Clevert et~al.(2015)Clevert, Unterthiner, and
  Hochreiter]{clevert2015fast}
Djork-Arn{\'e} Clevert, Thomas Unterthiner, and Sepp Hochreiter.
\newblock Fast and accurate deep network learning by exponential linear units
  (elus).
\newblock \emph{arXiv preprint arXiv:1511.07289}, 2015.

\bibitem[Courbariaux et~al.(2016)Courbariaux, Hubara, Soudry, El-Yaniv, and
  Bengio]{Courbariaux_2016_BinarizedNN}
Matthieu Courbariaux, Itay Hubara, Daniel Soudry, Ran El-Yaniv, and Yoshua
  Bengio.
\newblock Binarized neural networks: Training deep neural networks with weights
  and activations constrained to +1 or -1.
\newblock 2016.

\bibitem[Cybenko(1989)]{cybenko1989approximation}
George Cybenko.
\newblock Approximation by superpositions of a sigmoidal function.
\newblock \emph{Mathematics of control, signals and systems}, 2\penalty0
  (4):\penalty0 303--314, 1989.

\bibitem[Farhadi et~al.(2020)Farhadi, {Partovi Nia}, and
  Lodi]{farhadi2019activation}
Farnoush Farhadi, Vahid {Partovi Nia}, and Andrea Lodi.
\newblock Activation adaptation in neural networks.
\newblock Proceedings of the 9th International Conference on Pattern
  Recognition Applications and Methods (ICPRAM), to appear, 2020.

\bibitem[Gers et~al.(1999)Gers, Schmidhuber, and Cummins]{gers1999learning}
Felix~A Gers, J{\"u}rgen Schmidhuber, and Fred Cummins.
\newblock Learning to forget: Continual prediction with lstm.
\newblock 1999.

\bibitem[Glorot \& Bengio(2010)Glorot and Bengio]{glorot2010understanding}
Xavier Glorot and Yoshua Bengio.
\newblock Understanding the difficulty of training deep feedforward neural
  networks.
\newblock In \emph{Proceedings of the thirteenth international conference on
  artificial intelligence and statistics}, pp.\  249--256, 2010.

\bibitem[Glorot et~al.(2011)Glorot, Bordes, and Bengio]{glorot2011deep}
Xavier Glorot, Antoine Bordes, and Yoshua Bengio.
\newblock Deep sparse rectifier neural networks.
\newblock In \emph{Proceedings of the fourteenth international conference on
  artificial intelligence and statistics}, pp.\  315--323, 2011.

\bibitem[Griffiths \& Higham(2010)Griffiths and Higham]{Griffiths2010}
David~F. Griffiths and Desmond~J. Higham.
\newblock \emph{Euler's Method}, pp.\  19--31.
\newblock Springer London, London, 2010.
\newblock ISBN 978-0-85729-148-6.
\newblock \doi{10.1007/978-0-85729-148-6_2}.
\newblock URL \url{https://doi.org/10.1007/978-0-85729-148-6_2}.

\bibitem[Gulcehre et~al.(2016)Gulcehre, Moczulski, Denil, and
  Bengio]{gulcehre2016noisy}
Caglar Gulcehre, Marcin Moczulski, Misha Denil, and Yoshua Bengio.
\newblock Noisy activation functions.
\newblock In \emph{International conference on machine learning}, pp.\
  3059--3068, 2016.

\bibitem[He et~al.(2015)He, Zhang, Ren, and Sun]{he2015delving}
Kaiming He, Xiangyu Zhang, Shaoqing Ren, and Jian Sun.
\newblock Delving deep into rectifiers: Surpassing human-level performance on
  imagenet classification.
\newblock In \emph{Proceedings of the IEEE international conference on computer
  vision}, pp.\  1026--1034, 2015.

\bibitem[Hornik et~al.(1989)Hornik, Stinchcombe, White,
  et~al.]{hornik1989multilayer}
Kurt Hornik, Maxwell Stinchcombe, Halbert White, et~al.
\newblock Multilayer feedforward networks are universal approximators.
\newblock \emph{Neural networks}, 2\penalty0 (5):\penalty0 359--366, 1989.

\bibitem[Jozefowicz et~al.(2015)Jozefowicz, Zaremba, and
  Sutskever]{jozefowicz2015empirical}
Rafal Jozefowicz, Wojciech Zaremba, and Ilya Sutskever.
\newblock An empirical exploration of recurrent network architectures.
\newblock In \emph{International conference on machine learning}, pp.\
  2342--2350, 2015.

\bibitem[Krizhevsky et~al.(2012)Krizhevsky, Sutskever, and
  Hinton]{krizhevsky2012imagenet}
Alex Krizhevsky, Ilya Sutskever, and Geoffrey~E Hinton.
\newblock Imagenet classification with deep convolutional neural networks.
\newblock In \emph{Advances in neural information processing systems}, pp.\
  1097--1105, 2012.

\bibitem[Li et~al.(2015)Li, Karpathy, and Johnson]{li2015cs231n}
Fei-Fei Li, Andrej Karpathy, and Justin Johnson.
\newblock Cs231n: Convolutional neural networks for visual recognition.
\newblock \emph{University Lecture}, 2015.

\bibitem[Long et~al.(2017)Long, Lu, Ma, and Dong]{long2017pde}
Zichao Long, Yiping Lu, Xianzhong Ma, and Bin Dong.
\newblock Pde-net: Learning pdes from data.
\newblock \emph{arXiv preprint arXiv:1710.09668}, 2017.

\bibitem[Maas et~al.(2013)Maas, Hannun, and Ng]{maas2013rectifier}
Andrew~L Maas, Awni~Y Hannun, and Andrew~Y Ng.
\newblock Rectifier nonlinearities improve neural network acoustic models.
\newblock In \emph{Proc. icml}, volume~30, pp.\ ~3, 2013.

\bibitem[Montufar et~al.(2014)Montufar, Pascanu, Cho, and
  Bengio]{montufar2014number}
Guido~F Montufar, Razvan Pascanu, Kyunghyun Cho, and Yoshua Bengio.
\newblock On the number of linear regions of deep neural networks.
\newblock In \emph{Advances in neural information processing systems}, pp.\
  2924--2932, 2014.

\bibitem[Ramachandran et~al.(2017{\natexlab{a}})Ramachandran, Zoph, and
  Le]{Ramachandran_2017_SWISH}
Prajit Ramachandran, Barret Zoph, and Quoc~V Le.
\newblock Searching for activation functions.
\newblock \emph{arXiv preprint arXiv:1710.05941}, 2017{\natexlab{a}}.

\bibitem[Ramachandran et~al.(2017{\natexlab{b}})Ramachandran, Zoph, and
  Le]{ramachandran2017searching}
Prajit Ramachandran, Barret Zoph, and Quoc~V Le.
\newblock Searching for activation functions.
\newblock \emph{arXiv preprint arXiv:1710.05941}, 2017{\natexlab{b}}.

\bibitem[Sari et~al.(2019)Sari, Belbahri, and Nia]{sari2019study}
Eyy{\"u}b Sari, Mouloud Belbahri, and Vahid~Partovi Nia.
\newblock How does batch normalization help binary training?
\newblock \emph{arXiv preprint arXiv:1909.09139v2}, 2019.

\bibitem[van~den Berg(2016)]{van2016some}
Ewout van~den Berg.
\newblock Some insights into the geometry and training of neural networks.
\newblock \emph{arXiv preprint arXiv:1605.00329}, 2016.

\bibitem[Xu et~al.(2016)Xu, Huang, and Li]{xu2016revise}
Bing Xu, Ruitong Huang, and Mu~Li.
\newblock Revise saturated activation functions.
\newblock \emph{arXiv preprint arXiv:1602.05980}, 2016.

\end{thebibliography}

\end{document}